\documentclass[runningheads]{llncs}
\usepackage[T1]{fontenc}
\usepackage{graphicx}
\usepackage{subcaption}
\usepackage{amsfonts} 

\begin{document}
%
\title{Interpretable Deep Reinforcement Learning for Green Security Games with Real-Time Information}
\titlerunning{Interpretable DRL for GSG-I}
%
\author{Vishnu Dutt Sharma \and
John P. Dickerson \and
Pratap Tokekar}
%
%
\institute{University of Maryland, College Park MD 20740, USA
\email{\{vishnuds,john,tokekar\}@cs.umd.edu}}
\maketitle              
\begin{abstract}
 Green Security Games with real-time information (GSG-I) add the real-time information about the agents' movement to the typical GSG formulation. Prior works on GSG-I have used deep reinforcement learning (DRL) to
 learn the best policy for the agent in such an environment without any need to store the huge number of state representations for GSG-I. However, the decision-making process of DRL methods is largely opaque, which results in a lack of trust in their predictions. To tackle this issue, we present an interpretable DRL method for GSG-I that generates visualization to explain the decisions taken by the DRL algorithm. We also show that this approach performs better and works well with a simpler training regimen compared to the existing method.

\keywords{Green Security Games \and Deep Reinforcement Learning \and Interpretable Machine Learning}
\end{abstract}
\section{Introduction}
Green Security Games (GSGs) games are a sub-class of Stackelberg Security Games which are employed to model the interaction between law enforcement personnel and poachers in the green security domain like the security and conservation of forest, wildlife, and fisheries against poachers and illegal harvesters. Existing methods have used artificial intelligence for enhancing the movement strategies for law enforcement personnel, to capture the illegal and adversarial agents~\cite{klima2016markov}~\cite{basak2016combining}~\cite{kar2017ai}. Wang et al~\cite{wang2019deep} propose a variation of GSG called GSG-I, which includes the real-time footprints of the players. The addition of such information scales up the memory requirement to encode the game representation and makes the traditional solutions unsuitable for solving GSG-I. This issue can be resolved with deep reinforcement learning (DRL), a deep learning technique, treating the GSG as a Markov Decision Process (MDP). DRL can learn the policy or value function without supervision and does not need to store the state representation, unlike the tabular reinforcement learning approaches. The non-linear multi-layered structure of DRL networks makes them capable to learn the complex mapping between inputs and output, and therefore such networks can approximate the optimal policy well. Thus, DRL methods have been quite successful for pursuit-evasion games like Ms. Pacman~\cite{mnih2013playing}\cite{goldwaser2020deep}. Similarly, Wang et al.~\cite{wang2019deep} present a DRL method combined with the double oracle framework~\cite{mcmahan2003planning} called \textit{DeDOL} to GSG-I. The agent trained with this approach can use real-time information and performs well in terms of average episodic rewards.

While the complex structure of the DRL methods makes them successful in a variety of applications but also renders them opaque to humans i.e. decision-making processes of such methods are difficult to interpret. Real-life deployment of such systems can be dangerous in scenarios where the decision can cause harm to the end-user. Many recent works propose explainable frameworks and interpretable methods to allow the user to understand the key factors behind a decision and thus engender trust in DRL techniques. 

Motivated by these developments we present an interpretable DRL method to solve GSG-I. We also show that despite using interpretable constructs, this method performs better than the non-interpretable counterpart. Lastly, the training procedure for this method is simpler and faster than that of DeDOL.

\section{Related Work}
The idea of using DRL for GSGs has previously been explored for the continuous space GSGs~\cite{kamra2018policy}~\cite{kamra2019deepfp}, uncertainty augmented GSGs~\cite{venugopal2020reinforcement}, and Multi-agent games~\cite{lanctot2017unified}. The GSG definitions in these works lack real-time information that could capture the real-world interaction of agents with the environment. Wang et al.~\cite{wang2019deep} fill this gap by proposing GSG-I with the adversary's footprints as the real-time information. They also propose a DRL method trained with the double oracle framework~\cite{mcmahan2003planning}.  All of these methods, however, treat the DRL models as black-boxes, concealing their decision-making process. Making such systems transparent can help us ensure that the DRL policies do not result in any harm to the agents, and may provide further insights like robustness and generalizability of these systems. 

Efforts to bring transparency to the DRL models have gained momentum in the past few years. The approaches to make DRL, and deep learning in general, transparent can be classified into two directions: (a) interpretability, where the mechanism of the underlying model is analyzed to understand how a decision was reached, and (b) explainability, where we treat the DRL model as a black-box and try to approximate a simple, but interpretable method like decision tree over its prediction, to explain the reason behind an output. Interpretability is usually preferred over explainability as it tries to convey the actual decision-making process instead of approximating the original solution. But it is also often assumed to result in a worse-performing model, though no proof exists for this claim~\cite{rudin2021interpretable}.

As DRL networks have been used for a variety of domains, like computer vision and natural language processing, the existing work on transparent DRL doesn't have a general method to fit all the applications. While some focus on using the intermediate layers of a trained DRL model to visualize the feature influences~\cite{hilton2020understanding}~\cite{lewis2021deep}, others distill the knowledge into simpler prediction models and infer the decision rules from them~\cite{madumal2020explainable}. The states in GSG-I have spatial relations and can be thought of as multi-channel images. For such inputs, explainable approaches often employ saliency-based approaches~\cite{rosynski2020gradient}~\cite{atrey2019exploratory} are often used as they can be visualized along with the inputs. Tree-based methods like soft decision trees (SDTs)~\cite{dahlin2020designing} and Linear Model U-Trees (LMUTs)~\cite{liu2018toward} approximate the DRL model into a piece-wise linear function with its tree structure and visualize the importance of the input features.

For interpretability, the neural network architecture is required to have intermediate units whose output can be interpreted. The existing saliency-based methods~\cite{greydanus2018visualizing}\cite{fong2017interpretable}\cite{puri2019explain} rely on perturbations and thus result in explanations rather than interpretations. Object-based saliency maps~\cite{lewis2021deep} provide an alternative by using the appearance of the objects in the environment. But these maps also effectively generate explanations and rely heavily on domain knowledge. Hilton et al~\cite{hilton2020understanding} take a more rigorous approach of associating activations and data points to first identify the type of object that the neuron focuses on and then convert them to saliency maps. An simpler way to generate saliency maps is to use attention mechanism~\cite{bahdanau2014neural}. Attention mechanism is a neural network unit that generates the features by aggregating the input feature, where the weight for aggregation is learned using the same input features. The weights can be used to interpret the strength of influence of the features in the prediction. Prior works have successfully used attention mechanism for identifying the important part in images based on text for image captioning~\cite{xu2015show}, which supports their suitability for learning interpretable features in spatially-correlated inputs, or \textit{encoded images}. As the weights of attention mechanism participate in the decision-making process, it is an interpretable method. Attention-based saliency methods for DRL like Attention Augmented Agents ~\cite{liu2018toward} and Mask-Attention-A3C~\cite{itaya2021visual} generate their saliency maps by including spatial attention mechanism in the architecture. Mask-Attention-A3C uses a simpler implementation of attention mechanism akin to gates in the long short-term memory~\cite{hochreiter1997long} units and leverages them as masks to control the information flow form the feature maps to the classifier. This is a simpler yet effective implementation that can produce interpretable maps for both value and policy function, and thus we adapt it for solving GSG-I in this work.

\section{Experiment Setup}
The state in GSG-I is represented as a multi-dimensional array, encoding the information available to the patroller. However, Mask-Attention-A3C (MA-A3C) was originally designed for 3-channel color images. Thus we run experiments both on GSG-I states transformed to color images (referred to as \emph{color images} with the original MA-A3C architecture, and on multi-array representations (referred to as \emph{encoded images}) with a modified MA-A3C architecture. 

In the following subsections, we first provide details of the environment and then elaborate on the proposed DRL method.

\begin{figure}
     \centering
         \centering
        \includegraphics[width=\textwidth]{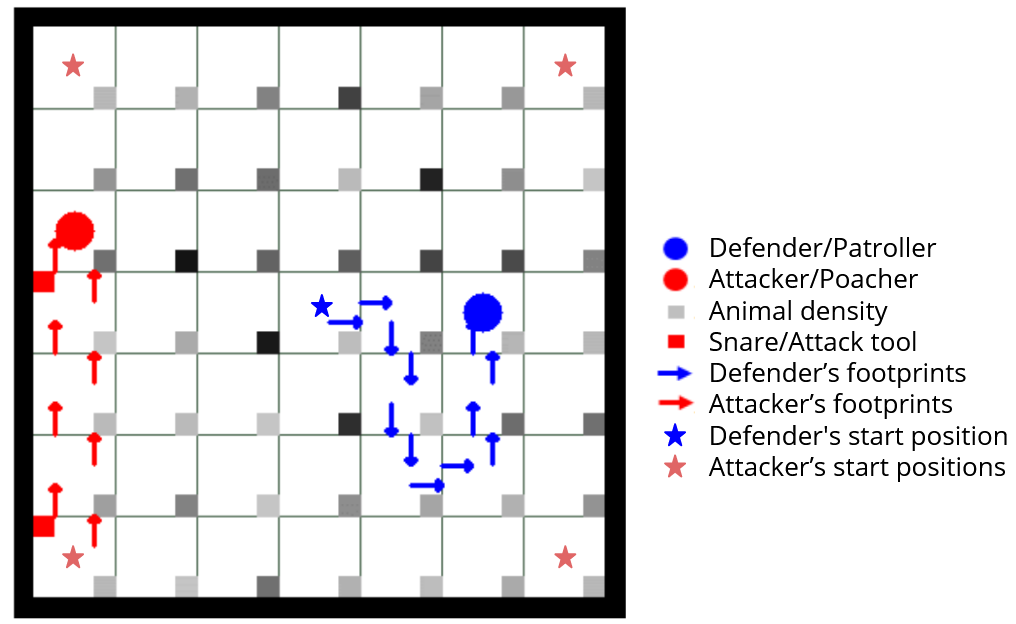}
    \caption{The GSG-I environment used in the experiments}
    \label{fig:gsg_intro}
\end{figure}


\subsection{GSG-I Environment}
Figure~\ref{fig:gsg_intro} shows a state from the GSG-I environment. This setting involves a patroller (defender) and a poacher (attacker). Each cell has a fixed probability of an animal being there, simulating areas in a forest, e.g. lake, mountain, swamp, etc. that differ in the fauna density. At the start of the game, the poacher, an adversary, enters the grid through one of the corners, chosen at random. Her goal is to capture animals by placing snares in the cells and exiting the environment through the starting location. A snare can launch a successful attack in a cell $(i,j)$ with probability $P^{attack}_{i,j}$ i.e. the probability of an animal's presence in that cell. She can place only 3 snares at max and tries to maximize the number of animals captured. Any attack launched after her return to the start position does not count. The patroller always starts patrolling from her post as the center of the grid. She has a map of the forest and thus knows the fauna density in each cell in the grid. The patroller can keep track of the real-time information like her and the attacker's footprints observed so far. The patroller's objective is to find and remove potential snares and catch the attacker as soon as possible.  The capture of the attacker results in the termination of the game. 

\begin{figure}
     \centering
     \begin{subfigure}[b]{0.45\textwidth}
         \centering
        \fbox{\includegraphics[width=\textwidth]{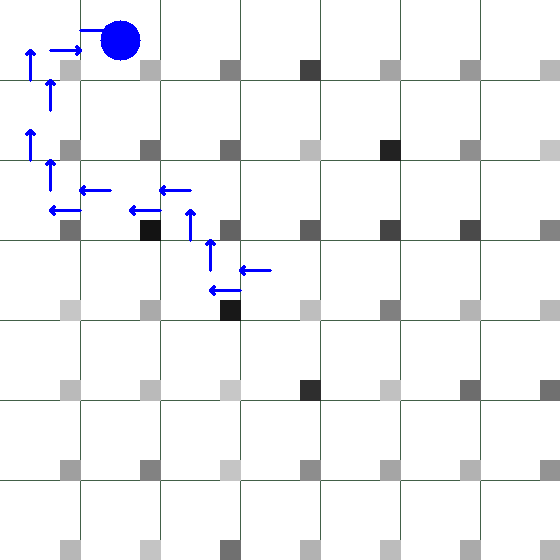}}
        \caption{State after $7^{th}$ step}
        \label{fig:gsg_interm_7}
    \end{subfigure}
    \hfill
    \begin{subfigure}[b]{0.45\textwidth}
         \centering
        \fbox{\includegraphics[width=\textwidth]{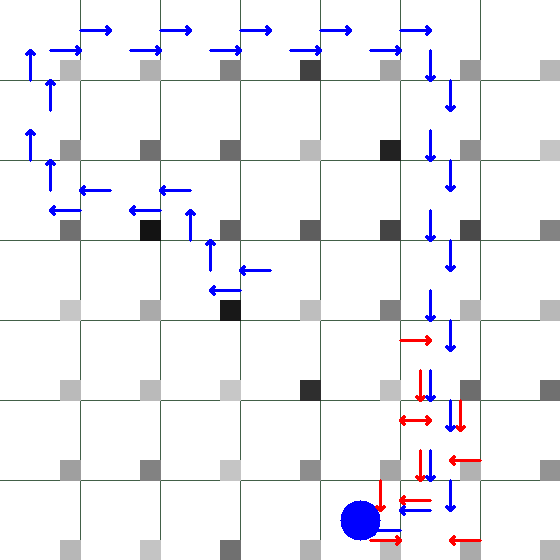}}
        \caption{State after $18^{th}$ step}
        \label{fig:gsg_interm_18}
    \end{subfigure}
    \caption{GSG-I states as images in an example episode at $7^{th}$ and $18^{th}$ state}
    \label{fig:gsg_img_state}
\end{figure}

GSG-I is designed as a two-player, zero-sum game. The patroller, our agent, gets a reward $\mathbb{R}$ of $+2$ when she removes a snare and a reward of $+8$ if she captures the attacker. But she incurs a negative reward equal to the $-P^{attack}_{i,j}$ if the attack is launched successfully in the cell $(i,j)$. The patroller can choose an action from her action space $\mathbb{A}^{d}$ : \textit{\{ up, down, right, left, stand still\}}. The attacker's actions space additionally includes placing snare as following, $\mathbb{A}^{a}$ : \textit{\{up, down, right, left, stand\, still \} $\times$ \{ place snare, do not place snare \} }. The game terminates if the patroller catches the attacker and finds all the snare, or if the time horizon $T$ is reached. The games state for encoded images $\mathbb{S}$ for the patroller, is represented 
 by a 20 channel square array with side 7, corresponding to the patroller's view of the environment and encodes the following information: the first 8 channels are the poachers' footprints encoded as binary variables indicating the direction of movement (\{north, west, east, south\}×\{entering or leaving\}); similarly, the next 8 channel encode the patroller's footprints; the $17^{th}$ channel encodes the patroller's location on the grid a 1 at the corresponding location in the array, and 0 elsewhere; the $18^{th}$ channel stores the animal density or the probability of successful attack by snare in each cell; the $19^{th}$ channel stores the trail of the patroller by adding 0.1 to a cell whenever the patroller visits it; the $20^{th}$ stores the time normalized by the time horizon $T$  (75 for the $7 \times 7$ grid GSG-I), which is same across all the cells.   The attacker can place a maximum of 6 snares in this grid. In our experiments, we use a randomly generated animal density map and attacker moves using heuristic random walk~\cite{wang2019deep}. The colored image representations are images of size $160 \times 160$ pixels generated with the information available to the patroller. Figure~\ref{fig:gsg_img_state} shows two colored images from an episode. As shown in Figure~\ref{fig:gsg_interm_18}, the patroller can not see the attacker and only knows about the attacker's footprints observed so far. 

\subsection{DRL Network}
\begin{figure}
     \centering
     \begin{subfigure}[b]{1.0\textwidth}
         \centering
        \includegraphics[width=\textwidth]{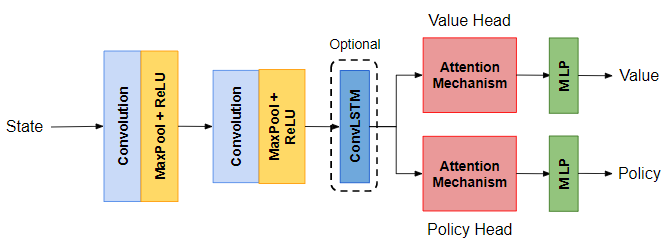}
        \caption{Network}
        \label{fig:maa3c_over}
    \end{subfigure}
    \hfill
    \begin{subfigure}[b]{0.70\textwidth}
         \centering
        \includegraphics[width=\textwidth]{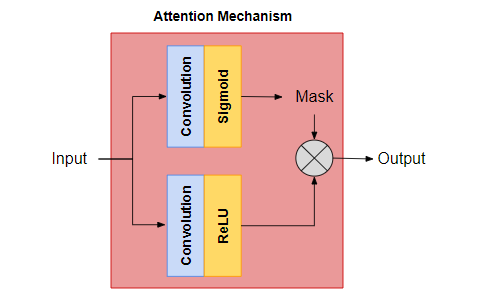}
        \caption{Attention mechanism for masking}
        \label{fig:maa3c_att}
    \end{subfigure}
    \caption{Mask-Attention-A3C Network Architecture}
    \label{fig:maa3c}
\end{figure}

MA-A3C network is composed of Convolution Neural Network (CNN) and Multi-Layer Perceptron (MLP) layers (Figure~\ref{fig:maa3c_over}). The CNN layers act as feature generators and for feature generation using the spatial relationship between entities. The feature generator for encoded images is composed of one convolutional and max-pooling, and for color images, we use 3 convolutional and max-pooling layers. After each max-pooling operation, ReLU activation is used. The output features maps have 64 channels and are shared between the policy (actor) and value (critic) heads. At each head, the network learns an attention map of the same size as the feature map. It is used as a mask over the output of the feature layer as shown in Figure~\ref{fig:maa3c_att}. The masks use sigmoid activation and thus act as almost binary filters controlling the flow of information; values close to 0 suppress the outputs at the corresponding location in the feature map, an values close to 1 pass them with minute change. The filters features are then fed to the MLP at each head. At policy head, the MLP acts as a regressor, whereas the value head acts as a classifier. The network additionally includes an optional ConvLSTM~\cite{shi2015convolutional} unit to share features across time and capture the temporal relations. We run experiments without ConvLSTM first and then enabled it on the better performing state representation to observe its effect. The experiments are conducted using a 32-core, 2.10Ghz Xeon Silver-4208 CPU and an Nvidia GeForce RTX 2080Ti GPU. We train the networks for ~100000 episodes, which takes around 18 hours for MA-A3C without ConvLSTM and around 27 hours for MA-A3C with ConvLSTM. The pre-trained DeDOL model, on the other hand, is reported to have been trained in local and global mode with a double oracle framework for 5 days. 

\section{Results}
\begin{figure}
     \centering
     \begin{subfigure}[b]{1.0\textwidth}
         \centering
        \includegraphics[width=\textwidth]{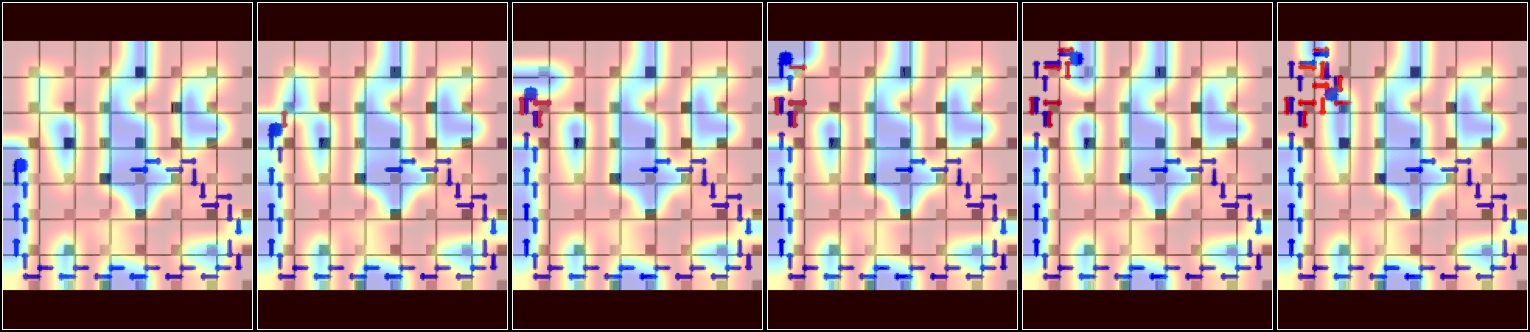}
        \caption{Attention maps for value head}
        \label{fig:gsg1_att_v}
    \end{subfigure}\\
    \begin{subfigure}[b]{1.0\textwidth}
         \centering
        \includegraphics[width=\textwidth]{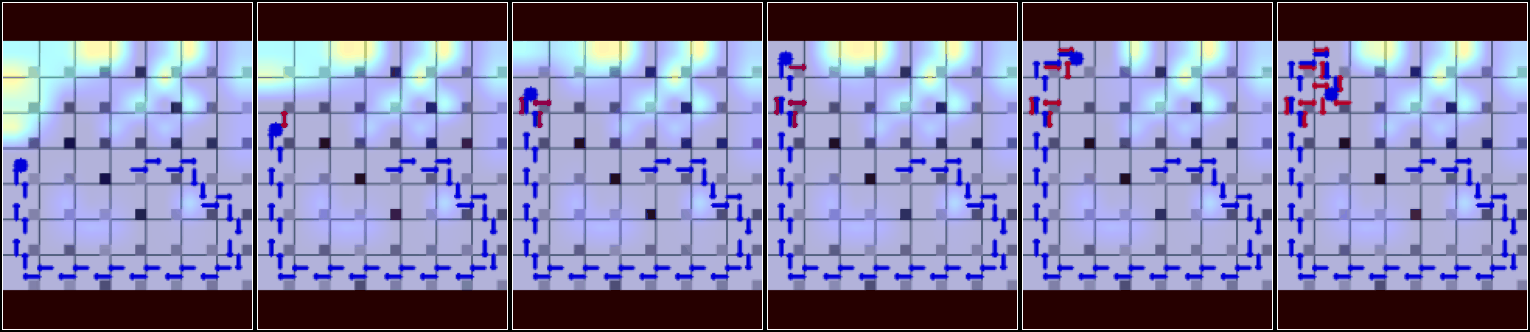}
        \caption{Attention maps for policy head}
        \label{fig:gsg1_att_p}
    \end{subfigure}
    \caption{Attention maps generated by Mask-Attention-A3C for GSG-I with 3-channel color image as the input}
    \label{fig:gsg1_mask_att}
\end{figure}

Our experiments show that the visualizing the attention maps/masks in MA-A3C provide some intuition about the next action.   Figure~\ref{fig:gsg1_mask_att} shows the examples for MA-A3C without ConvLSTM on color images. Here, the patroller moves towards the wall and follows along the edge of the environment until it comes across the attacker's footsteps. This behavior is similar to DeDOL's strategy presented by Wang et al~\cite{wang2019deep}. In the attention map for the value function, initially, the focus is on cells with high fauna density, which is reasonable as the agent hasn't collected any evidence about the attacker's movement. Also, the region of interest covers the patroller, a small part of the cell where she moves next, and her past few footprints. This is likely due to the focus on the new information that is available in the current cell. When multiple footprints of the attacker are seen, this map is useful in identifying where the patroller will move next. However, there are additional multiple regions on the map which do not seem to influence the patroller's actions. The attention map for the policy function seems less noisy compared to the value function, but it does not show the correct influential regions concerning the patroller's actions.

\begin{figure}
     \centering
     \begin{subfigure}[b]{1.0\textwidth}
         \centering
        \includegraphics[width=\textwidth]{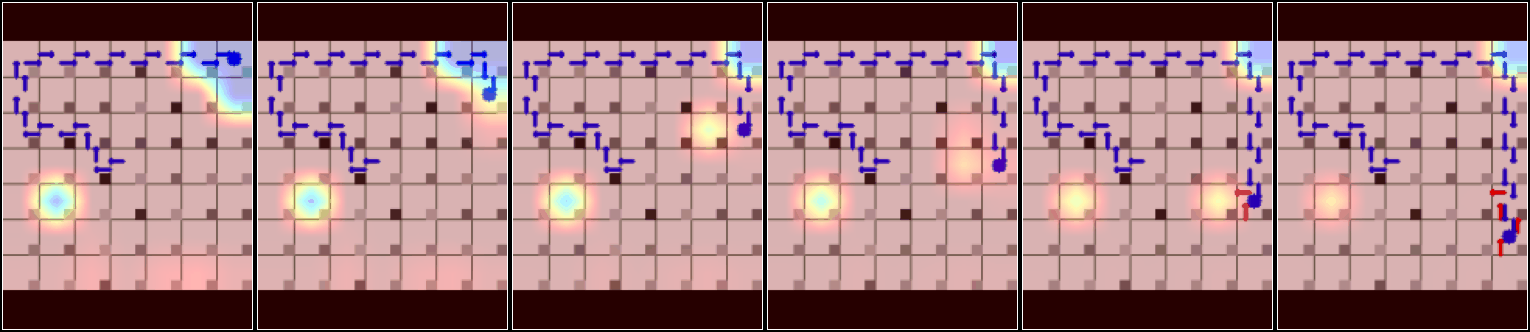}
        \caption{Attention maps for value head}
        \label{fig:gsg1_enc_att_v}
    \end{subfigure}\\
    \begin{subfigure}[b]{1.0\textwidth}
         \centering
        \includegraphics[width=\textwidth]{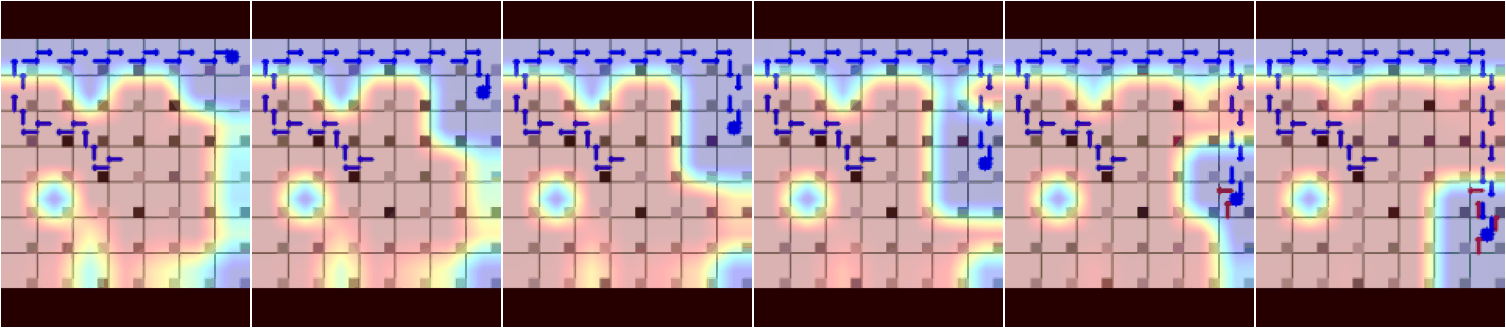}
        \caption{Attention maps for policy head}
        \label{fig:gsg1_enc_att_p}
    \end{subfigure}
    \caption{Attention maps generated by Mask-Attention-A3C for GSG-I with 20-channel encoded image as the input}
    \label{fig:gsg1_enc_mask_att}
\end{figure}

The interpretation maps of MA-A3C networks for the encoded images is comparatively more clear, as shown in Figure~\ref{fig:gsg1_enc_mask_att}. This was expected as the encoded images ease the burden of processing images into features for the neural network. In this case, the attention map for the value function is sparser and the area indicating the patroller's next step is not as salient here. On the other hand, the attention map for the policy function provides a better idea of the patroller's next steps. In the second last frame, the patroller could have moved left or down, but it is clear from the attention map that she will move down. An interesting observation here is that the initial path of the patroller is very similar to the path predicted by DeDOL (Figure~\ref{fig:gsg_interm_18}). Enabling ConvLSTM used results more intuitive and focused interpretable maps. As shown in Figure~\ref{fig:gsg1_enc_mask_lst}, both the value and policy maps give a clear sense of the agent's next move. We believe that this is because the agent is now able to compare the temporal changes in the state when making a decision and thus can process new information well. We also note that the attention maps in these experiments are not as clear and sharp as generally observed on games like Ms. Pacman, possibly due to local visibility of the agent in GSG-I as compared to Pacman, where the Pacman can see the whole environment.

\begin{figure}
     \centering
     \begin{subfigure}[b]{1.0\textwidth}
         \centering
        \includegraphics[width=\textwidth]{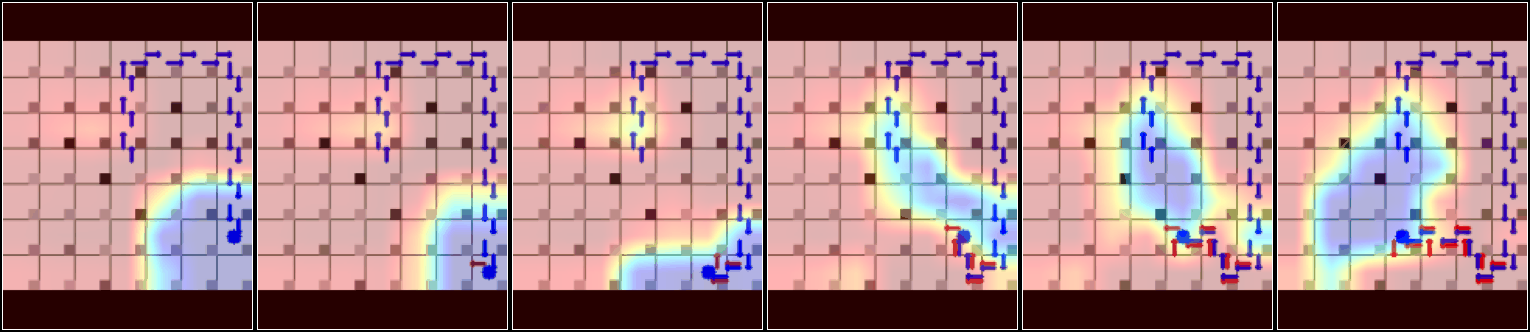}
        \caption{Attention maps for value head}
        \label{fig:gsg1_lstm_att_v}
    \end{subfigure}\\
    \begin{subfigure}[b]{1.0\textwidth}
         \centering
        \includegraphics[width=\textwidth]{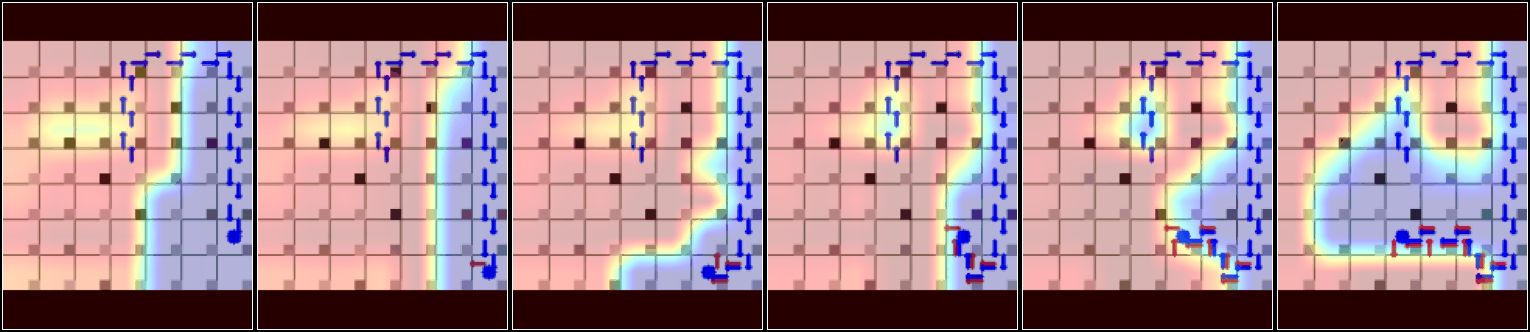}
        \caption{Attention maps for policy head}
        \label{fig:gsg1_lstm_att_p}
    \end{subfigure}
    \caption{Attention maps generated by Mask-Attention-A3C with ConvLSTM enabled for GSG-I with 20-channel encoded image as the input}
    \label{fig:gsg1_enc_mask_lstm}
\end{figure}

\begin{figure}
     \centering
     \begin{subfigure}[b]{0.9\textwidth}
         \centering
        \includegraphics[width=\textwidth]{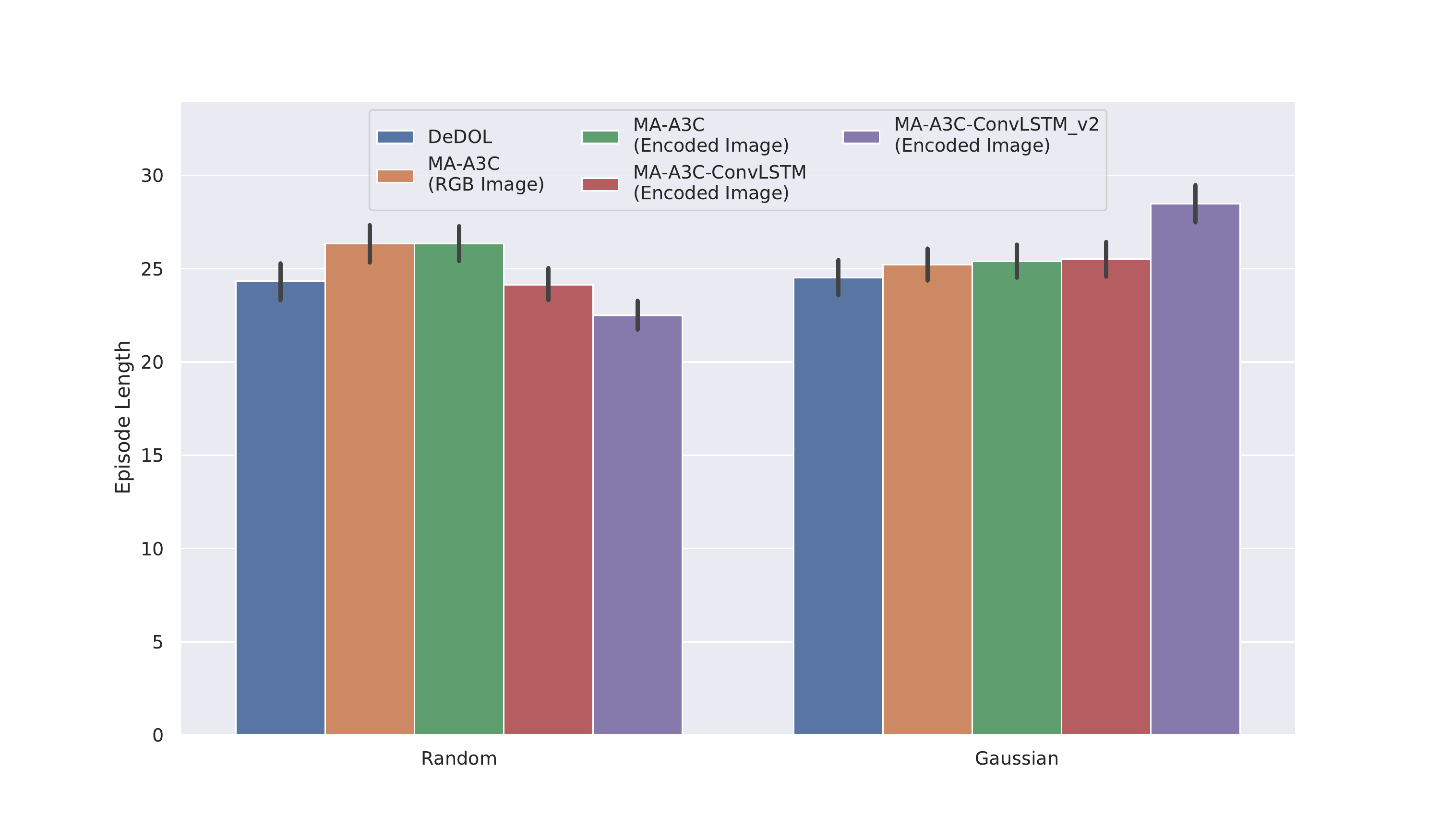}
        \caption{Episode length}
        \label{fig:eplen}
    \end{subfigure}\\
    \begin{subfigure}[b]{0.9\textwidth}
         \centering
        \includegraphics[width=\textwidth]{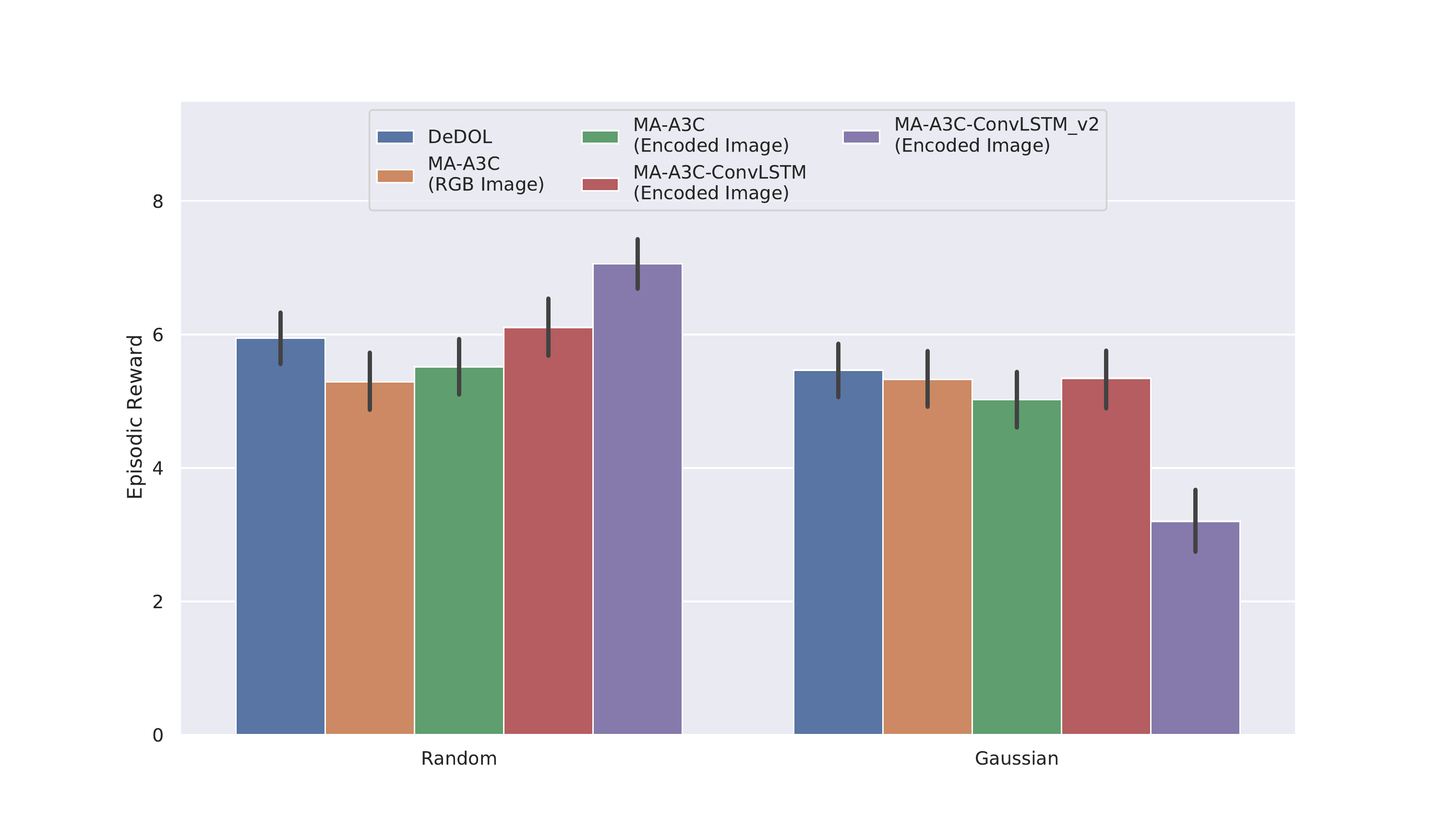}
        \caption{Episodic reward}
        \label{fig:epreward}
    \end{subfigure}
    \caption{Performance comparison of the interpretable DRL networks with DeDOL}
    \label{fig:epcomp}
\end{figure}

Figure~\ref{fig:epcomp} compares the episode length and the episodic reward i.e. the total reward accumulated in an episode, aggregated over 1000 episodes of GSG-I for DeDOL and the following variations of MA-A3C: MA-A3c over encoded image \textit{(MA-A3C (Encoded Image))}; MA-A3C over color images \textit{(MA-A3c (RGB Image))}; MA-A3c with ConvLSTM over encoded image \textit{(MA-A3C-ConvLSTM (Encoded Image))}; MA-A3C with ConvLSTM over encoded images trained for 1000,000 episodes \textit{(MA-A3C-ConvLSTM\_v2 (Encoded Image))}. We present these results for the randomly generated fauna density map and a Gaussian fauna density map (highest probability at the center and lower towards the edges). A good model should have low episode length (patroller catching the attacker as soon as possible) and high episodic reward (patroller removing all the snares and catching the attacker). We find the DeDOL to perform better than the MA-A3C models without ConvLSTMs but fares badly in comparison to ConvLSTM augmented MA-A3C models. While the \textit{MA-A3C-ConvLSTM (Encoded Image)} is able to perform on par with DeDOL, and further training (\textit{MA-A3C-ConvLSTM (Encoded Image)}) improves the margin of difference in metric from DeDOL. However, more training doesn't help the network generalize well and both the metric suffers severely compare to all other models. We suspect that the networks begin to memorize the input at this point. MA-A3C with the color image is similar to MA-A3C with the encoded image in performance. Thus we can simplify the network architecture if the preprocessed information is present, but raw information can also be used with appropriate architectural modifications.

\section{Conclusion and Future Work}
The experiments with Mask-Attention-A3C present promising qualitative and quantitative results. We found that similarly performing, but interpretable DRL models can be obtained by appropriate augmentations of attention maps and ConvLSTM units. Such models can be trained within lesser time and without complex training, regime consisting of multiple modes and game theoretic update mechanism. We believe that such methods can further improve with inclusion on units like ConvLSTM that indirectly act priors for learning methodology benefiting both the interpretation and performance.
An interesting direction for future work is the human evaluation of the map visualizations to conclude whether they are interpretable in general. A comparative evaluation of the two attention maps generated by Mask-Attention-A3C can add more insights. These maps provide two different interpretations. While an expert can relate them directly to the value function and the policy function, a non-expert may need the answer to the question \textit{which maps is more useful in a given condition}?. Such studies would therefore be able to generalize the meaning of \textit{interpretable visualizations}. 

\bibliographystyle{plain}
\bibliography{mybibliography}

\end{document}